\documentclass{article}

     \PassOptionsToPackage{numbers, compress}{natbib}
     \usepackage[preprint]{neurips_2019}

\usepackage[utf8]{inputenc} % allow utf-8 input
\usepackage[T1]{fontenc}    % use 8-bit T1 fonts
\usepackage{hyperref}       % hyperlinks
\usepackage{url}            % simple URL typesetting
\usepackage{booktabs}       % professional-quality tables
\usepackage{amsfonts}       % blackboard math symbols
\usepackage{nicefrac}       % compact symbols for 1/2, etc.
\usepackage{microtype}      % microtypography

\usepackage{multirow}
\usepackage{hhline}

\usepackage{times}
\usepackage{algorithm}
\usepackage{algorithmic}
\usepackage{lipsum}
\usepackage{epsfig}
\usepackage{amsmath}
\usepackage{amssymb}
\usepackage{graphicx} % more modern
\usepackage{subfigure}
\usepackage{epstopdf}
\usepackage{array}
\usepackage{bbm}

\def\p{\textbf{p}}
\def\s{\textbf{s}}

\def\w{\textbf{w}}
\def\x{\textbf{x}}
\def\y{\textbf{y}}

\def\R{\mathbb{R}}
\def\E{\mathbb{E}}
\def\X{\mathcal{X}}
\def\Y{\mathcal{Y}}
\def\Z{\mathcal{Z}}

\def\T{\top}

\title{Approaching Machine Learning Fairness through \\Adversarial Network}

\author{%
  Xiaoqian Wang$^1$, Heng Huang$^2$\\
  $^1$School of Electrical and Computer Engineering, Purdue University, USA \\
  $^2$Department of Electrical and Computer Engineering, Univeristy of Pittsburgh, USA \\
  \texttt{joy.xqwang@gmail.com, henghuanghh@gmail.com}
}

\begin{document}

\maketitle

\begin{abstract}
Fairness is becoming a rising concern \emph{w.r.t.} machine learning model performance. Especially for sensitive fields such as criminal justice and loan decision, eliminating the prediction discrimination towards a certain group of population (characterized by sensitive features like race and gender) is important for enhancing the trustworthiness of model. In this paper, we present a new general framework to improve machine learning fairness. The goal of our model is to minimize the influence of sensitive feature from the perspectives of both the data input and the predictive model. In order to achieve this goal, we reformulate the data input by removing the sensitive information and strengthen model fairness by minimizing the marginal contribution of the sensitive feature. We propose to learn the non-sensitive input via sampling among features and design an adversarial network to minimize the dependence between the reformulated input and the sensitive information. Extensive experiments on three benchmark datasets suggest that our model achieve better results than related state-of-the-art methods with respect to both fairness metrics and prediction performance.
%Through the backpropagation of sampling, our model automatically uncovers the features that are relevant to the sensitive information and train the model on the basis of non-sensitive features. This framework can be easily generalized to various machine learning models so as to improve model fairness while maintaining predictive performance. 
\end{abstract}

\section{Introduction}\label{introduction}

In recent years, machine learning has achieved unparalleled success in various fields, from face recognition, autonomous driving to computer-aided diagnosis. Despite the wide application and rapid development, the discrimination and bias that exists in machine learning models are attracting increasing attention in the research community. Recent models have been found to be biased towards certain groups of samples when making the prediction. For example, ProPublica \cite{propublica2016} analyzed a widely used criminal risk assessment tool for future crime prediction and discovered discrimination among different races. For defendants that do not commit a future crime, the black people are more likely to be mistaken by the model as potential future criminals than the white people (\emph{i.e.,} a higher false positive rate in the blacks than the whites). Moreover, Gross \emph{et al.} \cite{gross2014face} analyzed the face recognition problem and uncovered prediction discrimination among ethnicity, such that the recognition accuracy of white people is much higher than that of the black people.

Especially in sensitive fields such as criminal justice, credit and loan decision, and online advertising, a model with merely good prediction performance is not enough as we harness the power of machine learning. It is critical to guarantee that the prediction is based on appropriate information and the performance is not biased towards certain groups of population characterized by sensitive features like race and gender.

Improving model fairness is not only a societal problem but also an important aspect of machine learning. As the prediction bias uncovered in various applications, there are rising concerns \emph{w.r.t.} the discrimination among sensitive groups and thus the trustworthiness of model performance. In order to improve model fairness, recent works propose to achieve machine learning fairness from different perspectives. For example, as a pre-processing step, recent methods propose to eliminate the bias in data with reweighing the samples \cite{kamiran2012data} or removing the disparity among groups \cite{feldman2015certifying}. While in the in-processing of model prediction, Zhang \emph{et al.} \cite{zhang2018mitigating} proposes to improve fairness by constraining the prediction not based on sensitive information. Adel \emph{et al.} \cite{adel2019one} also propose an adversarial network that minimizes the influence of sensitive features to the prediction by characterizing the relevance between the latent data representation and the sensitive feature. What's more, fairness in prediction can be achieved with post-processing methods \cite{pleiss2017fairness} that modifies the model output for equalizing the probability of getting favorable output, \emph{e.g.,} getting approved for a loan.

Based on the targets of fairness, the motivation can be divided into group fairness and individual fairness. Group fairness is proposed to guarantee that different groups of population have equalized opportunity of achieving a favorable prediction result. Whereas for individual fairness \cite{zemel2013learning}, the goal is to guarantee that similar individuals get similar output. Based on the motivation of improving fairness, there are recent methods proposed to improve the long-term benefit of the protected groups (groups that are usually biased against by traditional models) \cite{liu2018delayed,mouzannar2019fair}, which is different than the methods that focus more on the instant benefit of an equalized opportunity \cite{pleiss2017fairness}.

Previous models usually propose to improve the fairness \emph{w.r.t.} either the data perspective or the model perspective, \emph{i.e.,} modifying the input to reduce data bias or optimizing the model to reduce prediction bias. These strategies may not guarantee the learned input to be optimal for the model or the designed model to be optimal for the data, such that a fairness constraint in the model usually introduces deterioration in the prediction performance.
%Moreover, previous works mainly focus on the sensitive feature but not the features that are not sensitive themselves but also contain sensitive information. % For example, \cite{pleiss2017fairness} propose to minimize the disparity in the prediction error among various groups of population via a post-processing step.

In order to improve fairness without sacrificing the predictive performance, we propose a new adversarial network to reduce the bias simultaneously from the data perspective and the model perspective. By conducting sampling among features, we automatically reformulate the input with features that contain only non-sensitive information. By minimizing the marginal contribution of the sensitive feature, we strengthen model robustness towards the sensitive feature such that adding sensitive information cannot influence the prediction results. The coupled optimization strategy from both the data and the model aspects improves fairness as well as prediction performance. We evaluate our model on three benchmark datasets, where our model achieves the best prediction performance as well as the most improved prediction fairness when compared with four state-of-the-art fairness models and the baseline.

The rest of the paper is organized as follows: in Section \ref{problem_def}, we introduce the terminologies in the paper and propose the motivation of our work. In Section \ref{model}, we propose our new model to improve fairness and derive the optimization algorithm. In Section \ref{experiments}, we conduct extensive experiments on three benchmark datasets and compare with five related methods using five different evaluation metrics to show the model performance in both prediction and fairness. Finally, we conclude the work in Section \ref{conclusion}.

\vspace{-8pt}
\section{Problem Definition}\label{problem_def}
We begin this section by introducing several terminologies in machine learning fairness.
%Let $\mathcal{Z}:=(\mathcal{X},\mathcal{Y})\subset \mathbb{R}^{d+1}$, where $\mathcal{X}\subset \mathbb{R}^d$ is a compact input space and $\mathcal{Y}$ is the set of labels. 

For a given dataset $[\x^{(1)},~\x^{(2)},~\dots,~\x^{(n)}]$ consisting of $n$ samples from the input space $\X\subset\R^d$, each sample $\x^{(i)} = [x^{(i)}_1,~x^{(i)}_2,~\dots,~x^{(i)}_d]^\T$ is characterized by $d$ features. In a prediction problem, \textbf{prediction bias} sometime exists when the model makes different prediction for different groups of samples with all other features held constant. For example, the Home Mortgage Disclosure Act (HMDA) data shows the rate of loan rejection is twice as high for blacks as for whites \cite{ladd1998evidence}.

The \textbf{sensitive feature} is the feature to characterize such \textbf{groups of population} of interest where we expect the prediction not to be biased among the groups. Examples of the sensitive feature include \emph{race, gender, age.} The choice of sensitive features varies for different prediction problems.

The \textbf{sensitive-relevant features} refers to the features that are not regarded as sensitive themselves, but indicate the information relevant to the sensitive feature. For example, in a job hiring decision model, the \emph{university} where the candidates graduate from is not a sensitive feature. However, \emph{university} can be relevant to the sensitive feature \emph{race} since the demographics of different universities is different.

One straightforward idea to improve fairness is \textbf{fairness through blindness}, \emph{i.e.,} simply exclude the sensitive feature from the input. However, this cannot eliminate the prediction bias, as the sensitive-relevant features still provide sensitive information in the input.

The goal of {fairness} varies in different applications, such as group/individual fairness, the long-term/instant benefit of fairness as introduced in Section \ref{introduction}. Here in this work, we are interested in improving the fairness with instant benefit among different {groups of population} so that the model prediction is not based on the sensitive information, either from the sensitive or sensitive-relevant features.

In this paper, we propose to reduce such prediction bias from two aspects: reformulating the \emph{input} and strengthening the  \emph{model} fairness. We achieve the goal by simultaneously learning a new input $\tilde\x$ based on the original data $\x$ and building a prediction model $f^\phi: \X \rightarrow \Y$ with the parameter $\phi$, where $\Y$ is the output space, such that 1) the dependency between $\tilde\x$ and the sensitive information is minimized; 2) the influence of the sensitive information to the prediction of $f^\phi$ is minimized. By improving from both the input and the model, we propose to guarantee that the prediction is based on the {non-sensitive information} and the bias \emph{w.r.t.} the sensitive feature is reduced.

\section{Approaching Machine Learning Fairness through Adversarial Network}\label{model}
\begin{figure}[!t]
	\centering
	\includegraphics[width=1\linewidth]{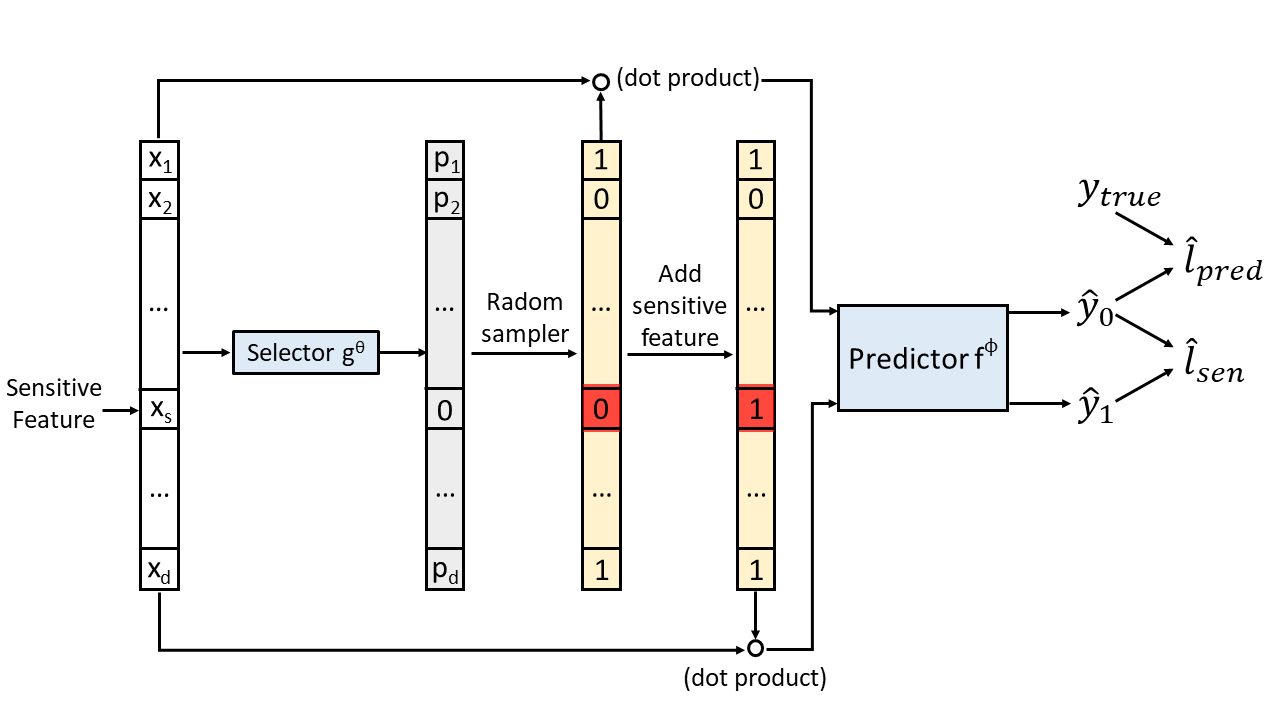}
	\caption{Illustration of the FAIAS model. FAIAS consists of a selector $g^\theta$ and a predictor $f^\phi$. The selector $g^\theta$ takes the feature vector as an input and predict the probability for each feature to be selected, based on which we randomly sample the features. The predictor $f^\phi$ gets two inputs, one (shown in the upper dot product) is the reformulated input using the sampled features, the other (shown in the bottom dot product) is by adding the sensitive feature to the sampled features. The difference between the output of $f^\phi$ \emph{w.r.t.} the two inputs is the sensitivity loss $\hat{l}_{sen}$, which shows the marginal contribution of the sensitive feature to the input. The prediction loss $\hat{l}_{pred}$ shows the prediction performance by using only sampled features.}
	\label{fig_illustration}
\end{figure}

As we discussed in Section \ref{problem_def}, the simple strategy of fairness through blindness cannot work with the existence of sensitive-relevant features. In order to reduce the prediction bias, we need to guarantee the prediction is not dependent on either the sensitive feature or the sensitive-relevant features. This is difficult to achieve since we usually do not have prior knowledge of what are the sensitive-relevant features. In this section, we propose a new FAIrness through AdverSarial network (FAIAS) model to improve the prediction fairness by improving both the input and the model.

The goal of reducing the prediction bias from both the input and model aspects can be formulated as two folds: 1) from the perspective of input, we propose to learn the new input $\tilde\x$ based on the original data $\x$ such that $\tilde\x$ contains only non-sensitive information; 2) for the prediction model, we minimize the marginal contribution of the sensitive feature such that adding the sensitive feature does not change the model prediction too much. 

We propose to learn the new input $\tilde\x$ by sampling the features in the original data $\x$, \emph{i.e.,} selecting features with a selection function $S: \X \rightarrow \{0,1\}^d$, such that the selected features contain only non-sensitive information.

For a data sample $\x = [x_1,~x_2,~\dots,~x_d]^\T \in\X$, and a selection set $\s = \{s_1,~s_2,~\dots,~s_m\} \subset \{1,~2,~\dots,~d\}$, denote $f^\phi(\x,\s) = f^\phi([x_{s_1},~x_{s_2},~\dots,~x_{s_m}])$ as the output of function $f^\phi$ when the input contains only features selected by $\s$ (the value of not selected features is set to $0$). For $t \not\in \s$, the marginal contribution of the $t$-th feature to this input can be denoted as $f^\phi(\x,\s\cup\{t\}) - f^\phi(\x,\s)$, \emph{i.e.,} the change in the output when adding $t$-th feature.

Denote the sensitive feature as $x_k$\footnote{For simplicity, here we only consider one sensitive feature in each data. It is notable that our FAIAS model can be easily applied to improving prediction fairness in the case involving multiple sensitive features.}, the goal of our FAIAS model is to minimize the marginal contribution $$f^\phi(\x,S\cup\{k\}) - f^\phi(\x,S),$$ where $S$ is the selection function that selects only features containing non-sensitive information.

We can approximate the selection function $S$ using a continuous selector function $g^\theta: \X \rightarrow [0,1]^d$ with parameter $\theta$, that takes the feature vector as the input and output a probability vector $\p = [p_1,~p_2,~\dots,~p_d]\in\R^d$ showing the probability to sample each feature to formulate the input. Then we conduct random sampling of the features based on the probability vector $\p$ and get the selection set $\s$. The probability of getting a joint selection vector $\s\in\{0,1\}^d$ is
\begin{eqnarray*}
%\label{pi}
\pi_\theta(\x,\s) = \Pi_{j=1}^d \big(g_j^\theta(\x)\big)^{s_j} \big(1-g_j^\theta(\x)\big)^{(1-s_j)}.
\end{eqnarray*}

To quantify the influence of sensitive feature, we consider the sensitivity loss as follows
\begin{eqnarray}
\label{lsen}
l_{sen}(\theta, \phi) = \E_{(\x,\y)\sim p}\E_{\s\sim \pi_\theta(\x,\cdot)} \Big[||f^{\phi}(\x, \s\cup\{k\}) - f^{\phi}(\x, \s)||\Big],
\end{eqnarray}
which characterize the marginal contribution of the sensitive feature $x_k$ to the model prediction given features selected by $\s$.
%In order to optimize the selector $g^\theta$ and the prediction model $f^\phi$, we consider two loss terms, sensitivity loss and prediction loss. 

In order to optimize $g^\theta$ to approximate the selection function $S$ and assign higher probability to only non-sensitive features, we propose an adversarial game between the selector function $g^\theta$ and the predictor function $f^\phi$.
%Define the selector function as $S: \X \rightarrow \{0,1\}^d$. 
%%\section{Approaching Group Fairness via Feature Contribution Learning}
%
%Define the selector function as $S: \X \rightarrow \{0,1\}^d$. We can approximate $S$ using a continuous function with parameter $\theta$ as $g^\theta: \X \rightarrow [0,1]^d$. The probability of getting a joint selection vector $\s\in\{0,1\}^d$ is
%\begin{eqnarray}
%\label{pi}
%\pi_\theta(\x,\s) = \Pi_{j=1}^d \big(g_j^\theta(\x)\big)^{s_j} \big(1-g_j^\theta(\x)\big)^{(1-s_j)}
%\end{eqnarray}
%where $f^{\phi}(\x, \s) = f^{\phi}(\x_\s)$.
%such that the model is able to make good prediction based on non-sensitive information. 

The goal of the prediction function $f^\phi$ is to minimize the sensitivity loss in Eq. \eqref{lsen} such that adding the sensitive feature does not influence the prediction too much. In contrast, we optimize the selector function $g^\theta$ to maximize the sensitivity loss in Eq. \eqref{lsen}, so as to select the subset of features which can be influenced the most by adding the sensitive feature. In this way, the selector function $g^\theta$ can find the features that are not relevant to the sensitive feature. If for example, the selected subset contains sensitive-relevant features, adding the sensitive feature will not bring too much change since the sensitive information is already indicated by the sensitive-relevant features. By updating the selector function $g^\theta$ to maximize the sensitivity loss, $g^\theta$ learns to exclude the sensitive information by assigning lower sampling probability to sensitive-relevant features and formulate the input on the basis of only non-sensitive information.

Moreover, we optimize the predictor $f^\phi$ to minimize the following prediction loss to guarantee prediction performance:
\begin{eqnarray}
\label{lpred}
l_{pred}(\theta, \phi) = \E_{(\x,\y)\sim p}\E_{\s\sim \pi_\theta(\x,\cdot)} \big[\sum\limits_{l=1}^c y_l \log f_l^{\phi}(\x, \s)\big],
\end{eqnarray}
which measures the performance of the prediction model given the features selected by $\s$. Here we take the the multi-class classification problem with $c$ class as an example and consider the cross entropy loss. We plot in Fig. \ref{fig_illustration} to show our FAIrness through AdverSarial network (FAIAS) model.

In order to optimize the selector function $g^\theta$ and the prediction function $f^\phi$, we derive the update steps for the two functions in the following.

Denote the empirical loss \emph{w.r.t} data $\x$ and selection vector $\s$ as below:
\begin{eqnarray*}
%\label{lsen_em}
%\hat l_{sen}(\x, \s) = ||f^{\phi}(\x, \s\cup\{k\}) - f^{\phi}(\x, \s)||
\hat l_{sen}(\x, \s) = f^{\phi}(\x, \s\cup\{k\}) - f^{\phi}(\x, \s),
%\hat l_{pred}(\x, \s) = \sum\limits_{l=1}^c y_l \log f_l^{\phi}(\x, \s),
\end{eqnarray*}
and the empirical loss as:
\begin{eqnarray*}
%\label{lpred_em}
\hat l_{pred}(\x, \s) = \sum\limits_{l=1}^c y_l \log f_l^{\phi}(\x, \s).
\end{eqnarray*}

The parameter $\theta$ and $\phi$ can be updated via gradient ascent and descent methods respectively. We can easily derive the derivative of $l_{sen}(\theta, \phi)$ \emph{w.r.t.} parameter $\theta$ and $\phi$ as follows:
\begin{eqnarray*}
\label{dsen_theta}
\nabla_\theta l_{sen}(\theta, \phi) &= &\nabla_\theta\E_{(\x,\y)\sim p}\E_{\s\sim \pi_\theta(\x,\cdot)} 
\Big[||f^{\phi}(\x, \s\cup\{k\}) - f^{\phi}(\x, \s)||\Big]\\
&=&\nabla_\theta\int_{\X\times\Y}p(\x,\y) \Big(\sum\limits_{\s\in\{0,1\}^d}\pi_\theta(\x,\s)
%||f^{\phi}(\x, \s\cup\{k\}) - f^{\phi}(\x, \s)||\\
||\hat l_{sen}(\x, \s)||\Big)dxdy\\
&=&\int_{\X\times\Y}p(\x,\y) \Big( \sum\limits_{\s\in\{0,1\}^d}\pi_\theta(\x,\s) \frac{\nabla_\theta\pi_\theta(\x,\s)}{\pi_\theta(\x,\s)}
%||f^{\phi}(\x, \s\cup\{k\}) - f^{\phi}(\x, \s)||\\
||\hat l_{sen}(\x, \s)||\Big)dxdy\\
&=&\int_{\X\times\Y}p(\x,\y) \Big( \sum\limits_{\s\in\{0,1\}^d}\pi_\theta(\x,\s) {\nabla_\theta \log\pi_\theta(\x,\s)}
%||f^{\phi}(\x, \s\cup\{k\}) - f^{\phi}(\x, \s)||\\
||\hat l_{sen}(\x, \s)||\Big)dxdy\\
&=&\E_{(\x,\y)\sim p}\E_{\s\sim \pi_\theta(\x,\cdot)} 
\Big[||\hat l_{sen}(\x, \s)|| \nabla_\theta \log\pi_\theta(\x,\s)\Big],
\end{eqnarray*}

\begin{eqnarray*}
\label{dsen_phi}
\nabla_\phi l_{sen}(\theta, \phi) &= &\nabla_\phi\E_{(\x,\y)\sim p}\E_{\s\sim \pi_\theta(\x,\cdot)} 
\big[||f^{\phi}(\x, \s\cup\{k\}) - f^{\phi}(\x, \s)||\big]\\
&= &\E_{(\x,\y)\sim p}\E_{\s\sim \pi_\theta(\x,\cdot)} 
\Big[\frac{(f^{\phi}(\x, \s\cup\{k\}) - f^{\phi}(\x, \s))(\nabla_\phi f^{\phi}(\x, \s\cup\{k\}) - \nabla_\phi f^{\phi}(\x, \s))}
%{||f^{\phi}(\x, \s\cup\{k\}) - f^{\phi}(\x, \s)||}\Big]
{||\hat l_{sen}(\x, \s)||}\Big],
\end{eqnarray*}
and the derivative of $l_{pred}(\theta, \phi)$ \emph{w.r.t.} $\phi$ is
\begin{eqnarray*}
\label{dpred_phi}
\nabla_\phi l_{pred}(\theta, \phi) &= &\nabla_\phi\E_{(\x,\y)\sim p}\E_{\s\sim \pi_\theta(\x,\cdot)} \big[\sum\limits_{l=1}^c y_l \log f_l^{\phi}(\x, \s)\big]\\
&= &\E_{(\x,\y)\sim p}\E_{\s\sim \pi_\theta(\x,\cdot)} \big[\sum\limits_{l=1}^c y_l \frac{\nabla_\phi f_l^{\phi}(\x, \s)}{f_l^{\phi}(\x, \s)}\big].
\end{eqnarray*}

In Algorithm \ref{alg} we summarize the optimization steps of FAIAS model. According to the update rules \emph{w.r.t.} the gradients, the time complexity of our FAIAS model is linear \emph{w.r.t.} the number of samples $n$, the number of parameters in $\theta$ and $\phi$, as well as the number of iterations $T$.
\begin{algorithm}[!t]
	\caption{Optimization Algorithm of FAIAS Model}
	\begin{algorithmic} \label{alg}
		\STATE \textbf{Input} data set $\Z = \X\times\Y = \{(\x_i,\y_i)\}_{i=1}^n$, learning rate $\alpha_\theta$ and $\alpha_\phi$. %hyper-parameter $\lambda$, 
		\STATE \textbf{Output} selector $g^\theta$ and predictor $f^\phi$.
		\STATE {Initialize} parameter $\theta$ and $\phi$ randomly. \;
		\WHILE{not converge}{
			\FOR{t = $1,~2,~\dots,~n_b$}{
				\FOR{$(\x_{t_i},\y_{t_i})$ in the $t$-th mini-batch $\Z_t$}{
					\STATE 1. Calculate the selection probability vector $$g^\theta(\x_{t_i}) = [p_{t_i}^1,~p_{t_i}^2,~\dots,~p_{t_i}^d].$$\;
					\STATE 2. Sample the selection vector $\s_{t_i}\in\R^d$ with $$\s_{t_i}^j\sim Bernoulli(p_{t_i}^j), \quad\text{for} ~j = 1,~,2,~\dots,~d.$$\;
					%for $j=1,~2,~\dots,~d$.\;
					\STATE 3. Calculate $$\hat l_{sen}(\x_{t_i}, \s_{t_i}) = f^{\phi}(\x_{t_i}, \s_{t_i}\cup\{k\}) - f^{\phi}(\x_{t_i}, \s_{t_i}).$$\;
				}\ENDFOR
				\STATE 4. Update the parameter $\theta$ with gradient ascent
				$$ \theta \leftarrow \theta +  \frac{\alpha_\theta}{n_b} \sum\limits_{i}^{}\hat l_{sen}(\x_{t_i}, \s_{t_i}) \nabla_\theta \log\pi_\theta(\x_{t_i},\s_{t_i}).$$\;
				\STATE 5. Update the parameter $\phi$ with gradient descent
%				$$\phi \leftarrow \phi - \alpha_\phi\frac{\hat l_{sen}(\x_{t_i}, \s_{t_i})\big(\nabla_\phi f^{\phi}(\x_{t_i}, \s_{t_i}\cup\{k\}) - \nabla_\phi f^{\phi}(\x_{t_i}, \s_{t_i})\big)}
				$$\phi \leftarrow \phi - \frac{\alpha_\phi}{n_b} \sum\limits_{i}^{} \frac{\hat l_{sen}(\x_{t_i}, \s_{t_i})\nabla_\phi \hat l_{sen}(\x_{t_i}, \s_{t_i})}
					{||\hat l_{sen}(\x_{t_i}, \s_{t_i})||}
					-  \frac{\alpha_\phi}{n_b} \sum\limits_{i}^{} \sum\limits_{l=1}^c y_l \frac{\nabla_\phi f_l^{\phi}(\x_{t_i}, \s_{t_i})}{f_l^{\phi}(\x_{t_i}, \s_{t_i})}.$$\;
			}\ENDFOR
		}
		\ENDWHILE
	\end{algorithmic}
\end{algorithm}

%\textbf{Time Complexity}:

\section{Experimental Results}\label{experiments}

\begin{figure}[!t]
	\centering
	\includegraphics[width=0.42\linewidth]{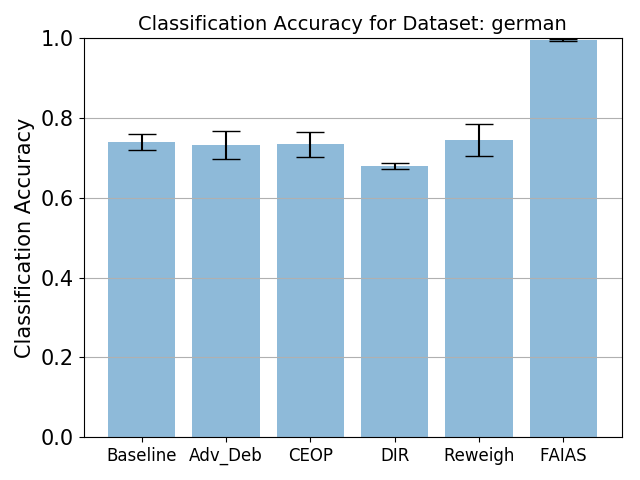}
	\quad\quad
	\includegraphics[width=0.42\linewidth]{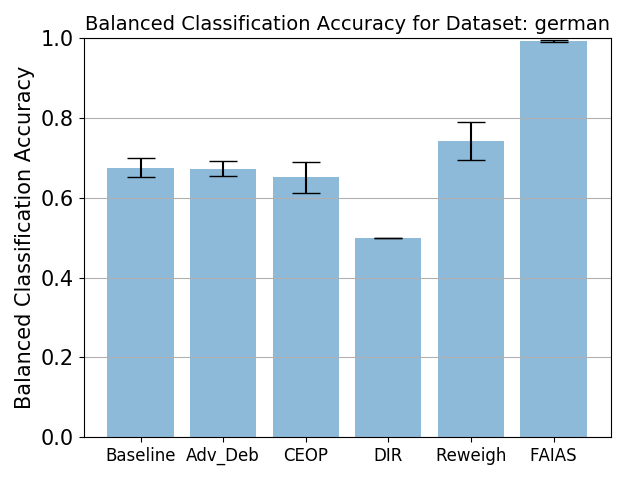}
	\includegraphics[width=0.42\linewidth]{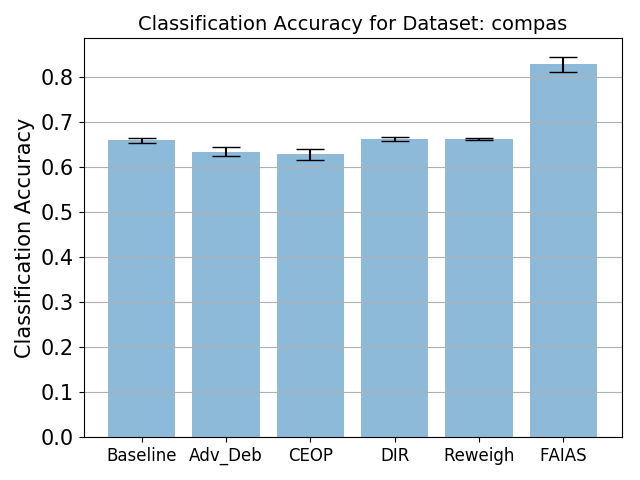}
	\quad\quad
	\includegraphics[width=0.42\linewidth]{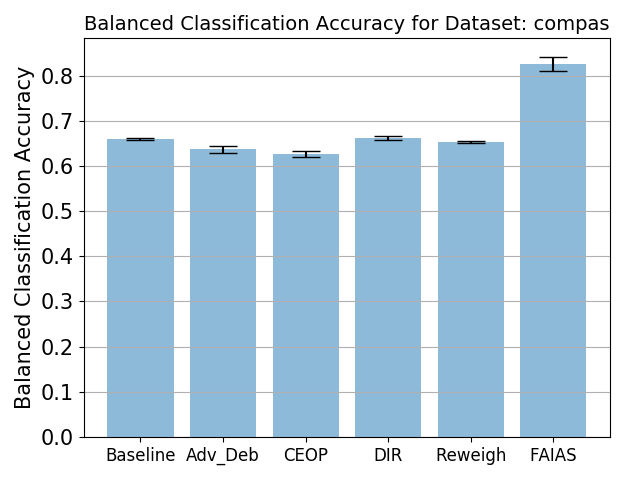}
	\includegraphics[width=0.42\linewidth]{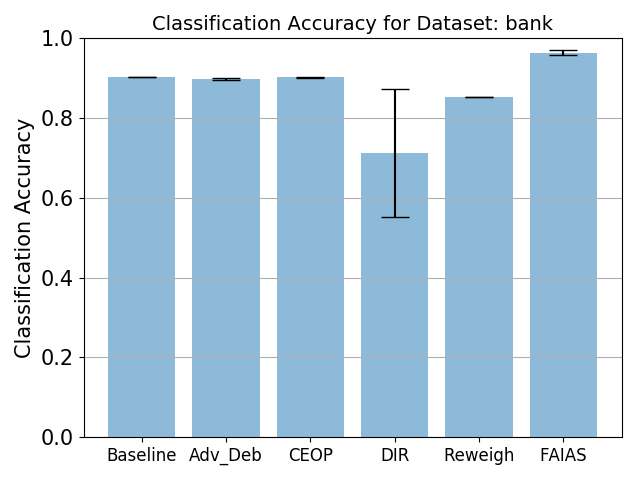}
	\quad\quad
	\includegraphics[width=0.42\linewidth]{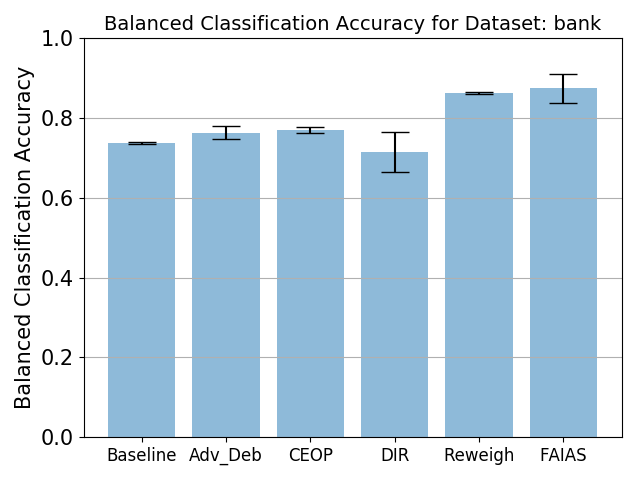}
	\caption{Comparison of model performance via classification accuracy and balanced classification accuracy on three benchmark datasets.}
	\label{fig_acc}
\end{figure}

In this section, we conduct experiments on three benchmark datasets to validate the performance of our FAIAS model. It is notable that our FAIAS model is proposed for group fairness, \emph{i.e.,} minimizing the prediction bias \emph{w.r.t.} a certain sensitive feature in both the pre-processing and in-processing steps. We compare our model with four recent methods for group fairness in pre-processing, in-processing, and post-processing steps, and one baseline method as follows.
\begin{itemize}
	\item \textbf{Baseline method without fairness constraint}: the logistic regression model that adopts all features (including the sensitive feature) in the training and prediction.
	\item \textbf{Adversarial de-biasing model} (abbreviated as Adv$\_$Deb in the comparison)\cite{zhang2018mitigating}: an in-processing model that proposes to maximize the predictive performance while minimizing the adversary's ability to predict the sensitive features.
	\item \textbf{Calibrated equal odds post-processing} (abbreviated as CEOP in the comparison)\cite{pleiss2017fairness}: a post-processing model that proposes to minimize the error disparity among different groups indicated by the sensitive feature.
	\item \textbf{Disparate impact remover} (abbreviated as DIR in the comparison) \cite{feldman2015certifying}: a model that proposes to minimize the disparity in the outcome from different groups via pre-processing.
	\item \textbf{Reweighing method} \cite{kamiran2012data}: a pre-processing method that eliminates the discrimination bias among different groups by reweighing and resampling the data.
\end{itemize}

We use three benchmark datasets to compare the model performance:% \footnote{In order to compare with the relevant methods, here we only consider one sensitive feature in each data. It is notable that our FAIAS model can be easily applied to improving model fairness in the case involving multiple sensitive features.}:
\begin{itemize}
	\item \textbf{German credit data} from the UCI repository \cite{Dua2019}: The data contains 1000 samples described by 20 features and the goal is to predict the credit risks (good or bad). The feature \emph{personal status and sex} is used as the sensitive feature.
	\item \textbf{Compas}\footnote{Compas data is downloaded from \url{https://github.com/propublica/compas-analysis}}: The data includes 6167 samples described by 401 features with the outcome showing if each person was accused of a crime within two years. The feature \emph{gender} is used as the sensitive feature in this data.
	\item \textbf{Bank marketing data}\cite{speicher2018unified} from the UCI repository: The data consists of 45211 samples with 17 features. The goal is to predict whether a client will subscribe to a term deposit. The sensitive feature in this data is \emph{age}.
\end{itemize}

We use the \textbf{classification accuracy} (percentage of correctly classified data in the testing set) and \textbf{balanced classification accuracy} (average of true positive rate and true negative rate) to evaluate the model prediction performance in the classification problem. Moreover, we adopt three different metrics to evaluate the fairness among groups of population \emph{w.r.t.} the sensitive feature in the data:
\begin{itemize}
	\item \textbf{Absolute equal opportunity difference}: the absolute difference in true positive rate among different groups of population.
	\item \textbf{Absolute average odds difference}: the absolute difference in balanced classification accuracy among different groups of population.
	\item \textbf{Theil index}: proposed in \cite{speicher2018unified} to measure the group or individual fairness. Here we report the absolute value of the Theil index, which is always positive. A close-to-zero Theil index indicates more fairness.
\end{itemize}

Features in the data are normalized to the range of $[0,~1]$. For each dataset, we randomly split the data into three sets: $60\%$ for training set, $20\%$ for validation set, and $20\%$ for testing set, where the training set is used to train the model, validation set is used to tune the hyper-parameter, and the test is used to test the model performance. We run all comparing methods $5$ times with $5$ different random splits of the data and report the average performance with the standard deviation on the test set. For the methods involving a hyper-parameter, \emph{i.e.,} the thresholding value in CEOP, DIR, and Reweighing method, we tune the hyper-parameter in the range of $\{0,~0.1,~0.2,~\dots,~1\}$ and use the best hyper-parameter achieving the best balanced classification accuracy on the validation set.

We implement the comparing methods via the AI Fairness 360 toolbox \cite{bellamy2018ai}. For our FAIAS model, we construct the predictor as a $4$-layer neural network with 200 nodes in each layer. We adopt scaled exponential linear units (SELU)\cite{klambauer2017self} as the activation function of the first $3$ layers and the softmax function for the last layer. We use Adam optimizer \cite{kingma2014adam} and set the learning rate as $0.0001$. For the selector, we set it as a data-independent vector $\w = \frac{1}{1+e^{-\theta}} \in\R^d$ since we expect the selected features to be consistent among different samples. We use Tensorflow and Keras toolbox for implementing our code and run the algorithm on a machine with one Titan X Pascal GPU.

%All experiments are conducted on a 72-core Intel(R) Xeon(R) E5-2699 v3 CPU @ 2.30GHz server with 528GB memory. The operating system is Ubuntu 16.04.1, and the software is Matlab R2016a (64-bit) 9.0.0.

\begin{figure}[!t]
	\centering
	\includegraphics[width=0.32\linewidth]{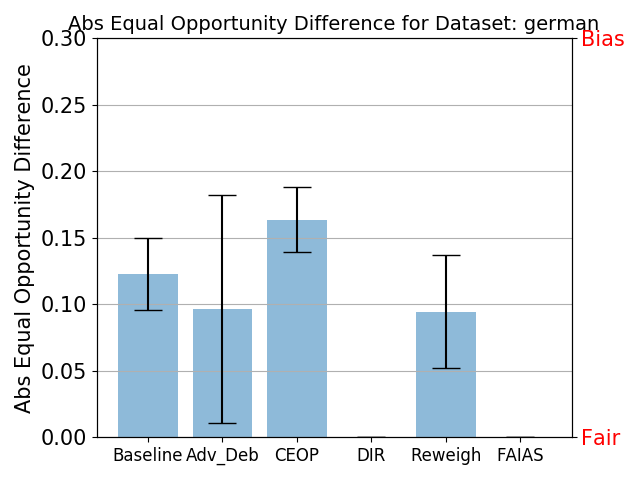}
	\includegraphics[width=0.32\linewidth]{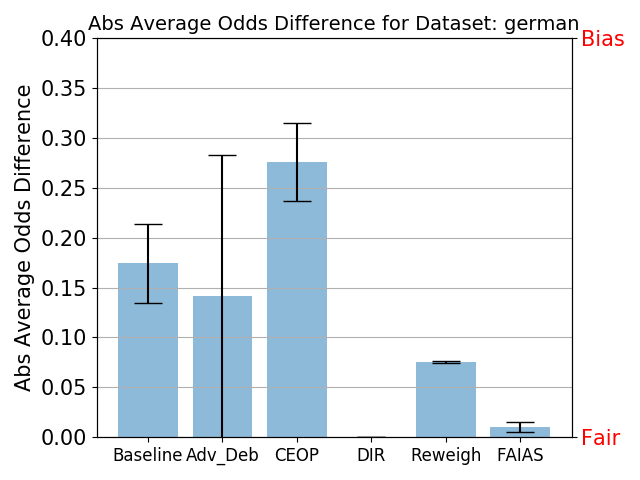}
	\includegraphics[width=0.32\linewidth]{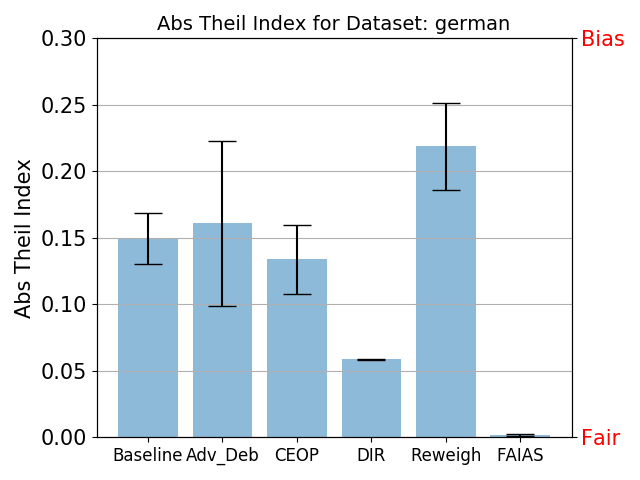}	
	
	\includegraphics[width=0.32\linewidth]{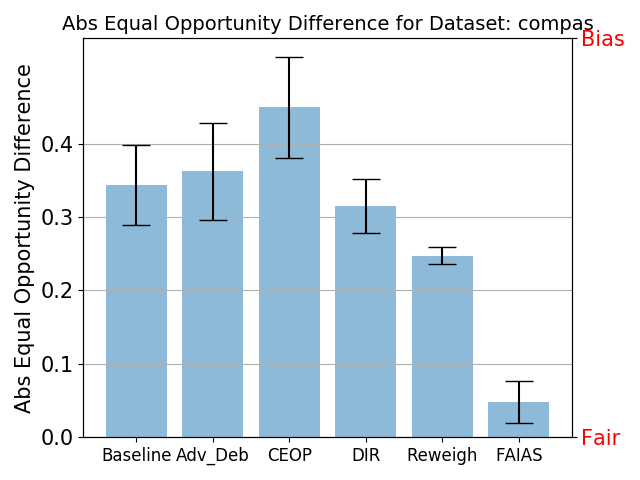}
	\includegraphics[width=0.32\linewidth]{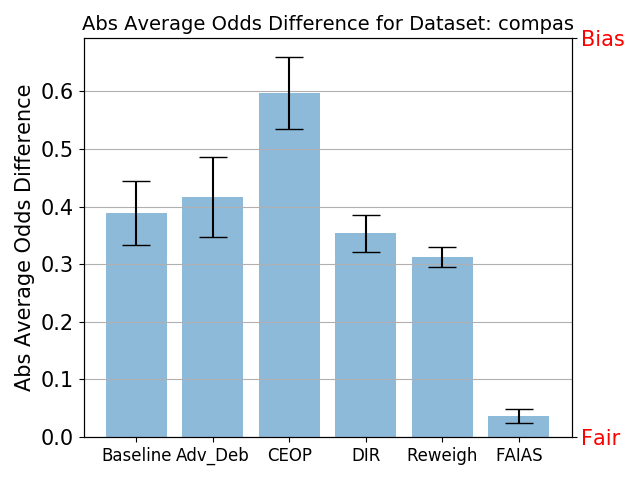}
	\includegraphics[width=0.32\linewidth]{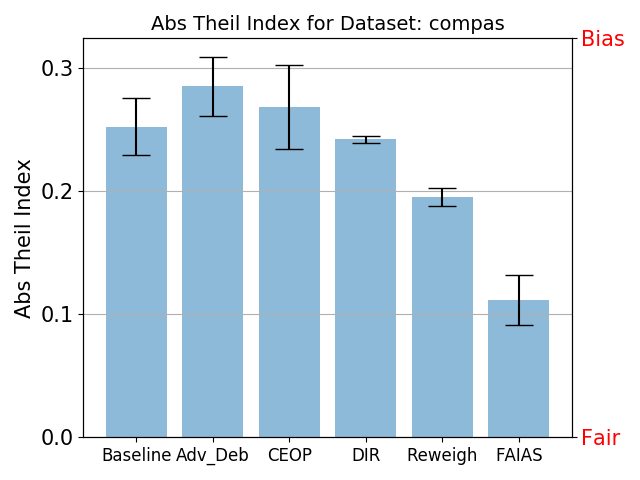}
	
	\includegraphics[width=0.32\linewidth]{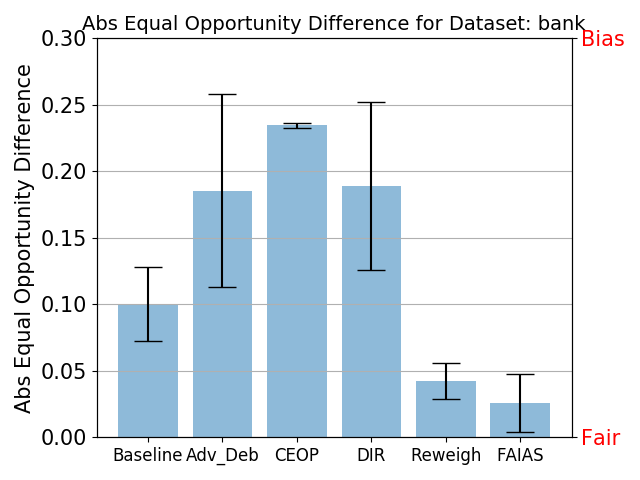}
	\includegraphics[width=0.32\linewidth]{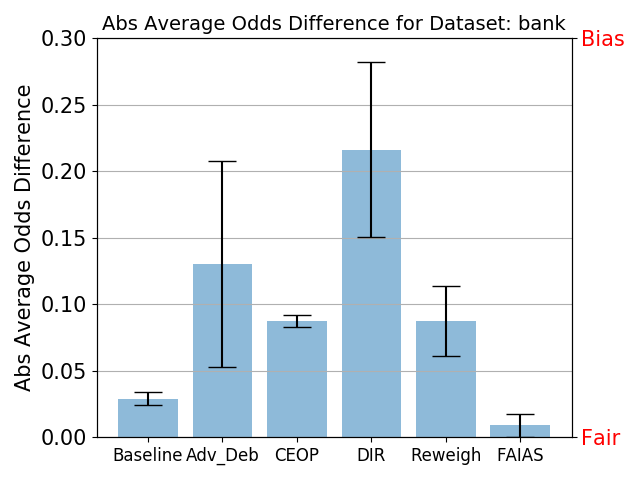}
	\includegraphics[width=0.32\linewidth]{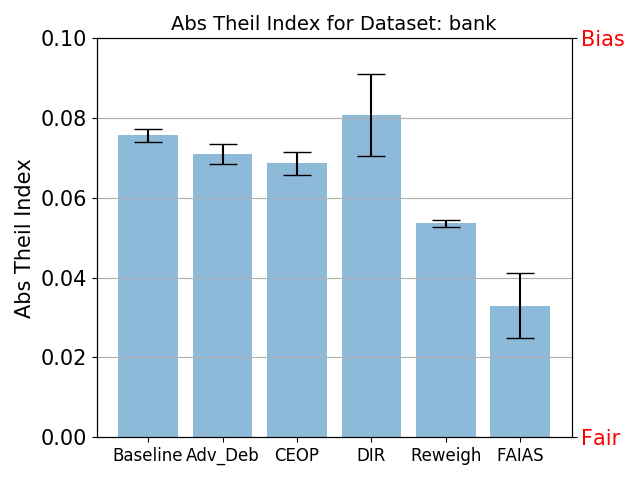}
	\caption{Comparison of prediction fairness via absolute equal opportunity difference, absolute average odds difference, and Theil index on three benchmark datasets.}
	\label{fig_fair}
\end{figure}

We first compare the model performance on the classification problems and summarize the results in Fig. \ref{fig_acc}. The results show that our FAIAS model achieves the best classification result \emph{w.r.t.} both the accuracy and the balanced accuracy, which indicate that the optimization on both the data and model perspective is successful in guaranteeing the prediction performance such that imposing the fairness constraints does not sacrifice the classification performance.

We also use the three fairness metrics to evaluate if FAIAS improves the prediction fairness by rendering equal prediction performance among different groups of population. From the results in Fig. \ref{fig_fair}, we can notice that FAIAS achieves equivalent or better results \emph{w.r.t.} all three measurement metrics on the three benchmark datasets, such that the feature sampling via an adversarial network is able to eliminate the sensitive information and forces the prediction performance to be equalized among different groups of population.

\section{Conclusion}\label{conclusion}
In this work, we propose a new adversarial network FAIAS for improving prediction fairness. We formulate our model from both the data perspective and the model perspective. Our FAIAS model consists of two components: a selector function and a prediction function, where the selector function is optimized on the data perspective to select the features containing only non-sensitive information and the prediction function is optimized from the model perspective to minimize the marginal contribution of the sensitive feature and also improve the prediction performance. We conduct extensive experiments on three benchmark datasets and validate that our FAIAS model outperforms all related methods \emph{w.r.t.} both the prediction performance and fairness metrics.

%\subsubsection*{Acknowledgments}
%
%Use unnumbered third level headings for the acknowledgments. All acknowledgments
%go at the end of the paper. Do not include acknowledgments in the anonymized
%submission, only in the final paper.

%\section*{References}
%
%References follow the acknowledgments. Use unnumbered first-level heading for
%the references. Any choice of citation style is acceptable as long as you are
%consistent. It is permissible to reduce the font size to \verb+small+ (9 point)
%when listing the references. {\bf Remember that you can use more than eight
%  pages as long as the additional pages contain \emph{only} cited references.}
%\medskip
%
%\small
%
%[1] Alexander, J.A.\ \& Mozer, M.C.\ (1995) Template-based algorithms for
%connectionist rule extraction. In G.\ Tesauro, D.S.\ Touretzky and T.K.\ Leen
%(eds.), {\it Advances in Neural Information Processing Systems 7},
%pp.\ 609--616. Cambridge, MA: MIT Press.
%
%[2] Bower, J.M.\ \& Beeman, D.\ (1995) {\it The Book of GENESIS: Exploring
%  Realistic Neural Models with the GEneral NEural SImulation System.}  New York:
%TELOS/Springer--Verlag.
%
%[3] Hasselmo, M.E., Schnell, E.\ \& Barkai, E.\ (1995) Dynamics of learning and
%recall at excitatory recurrent synapses and cholinergic modulation in rat
%hippocampal region CA3. {\it Journal of Neuroscience} {\bf 15}(7):5249-5262.

%\bibliography{nips19}
%\bibliography{nips17}
%\bibliographystyle{abbrv}

\end{document}